\documentclass[sn-mathphys-num]{sn-jnl}

\usepackage{multirow}
\usepackage{booktabs}
\usepackage{graphicx}%
\usepackage{multirow}%
\usepackage{float}
\usepackage{amsmath,amssymb,amsfonts}%
\usepackage{amsthm}%
\usepackage{mathrsfs}%
\usepackage[title]{appendix}%
\usepackage{xcolor}%
\usepackage{textcomp}%
\usepackage{hyperref}%
\usepackage{xr-hyper}%
\usepackage{manyfoot}%
\usepackage{booktabs}%
\usepackage{algorithm}%
\usepackage{algorithmicx}%
\usepackage{algpseudocode}%
\usepackage{listings}%
\usepackage{multirow}
\usepackage{booktabs}
\usepackage{tabularx}
\usepackage{caption} 
\usepackage{longtable}

\theoremstyle{thmstyleone}%

\theoremstyle{thmstyletwo}%

\theoremstyle{thmstylethree}%

\begin{document}

\title[Article Title]{A Comparative Investigation of Thermodynamic Structure-Informed Neural Networks}


\author[1]{\fnm{Guojie}\sur{Li}}

\author*[1]{\fnm{Liu}\sur{Hong}}\email{hongliu@sysu.edu.cn}
\affil[1]{\orgdiv{School of Mathematics}, \orgname{Sun Yat-sen University}, \orgaddress{\city{Guangzhou}, \postcode{510275}, \country{China}}}

\abstract{Physics-informed neural networks (PINNs) offer a unified framework for solving both forward and inverse problems of differential equations, yet their performance and physical consistency strongly depend on how governing laws are incorporated. In this work, we present a systematic comparison of different thermodynamic structure-informed neural networks by incorporating various thermodynamics formulations, including Newtonian, Lagrangian, and Hamiltonian mechanics for conservative systems, as well as the Onsager variational principle and extended irreversible thermodynamics for dissipative systems. Through comprehensive numerical experiments on representative ordinary and partial differential equations, we quantitatively evaluate the impact of these formulations on accuracy, physical consistency, noise robustness, and interpretability. The results show that Newtonian-residual-based PINNs can reconstruct system states but fail to reliably recover key physical and thermodynamic quantities, whereas structure-preserving formulation significantly enhances parameter identification, thermodynamic consistency, and robustness. These findings provide practical guidance for principled design of thermodynamics-consistency model, and lay the groundwork for integrating more general nonequilibrium thermodynamic structures into physics-informed machine learning.}




\maketitle

\section{Introduction}

Learning dynamical systems whose solutions are consistent with data is an important task in scientific machine learning and has received widespread attention \cite{wan2018data, ghnatios2019data, karapiperis2021data}. One notable category of works is the physics-informed machine learning (PIML). The seminal contribution on this subject dates back to the earlier study of Lagaris \cite{lagaris1998artificial} and Owhadi \cite{owhadi2015bayesian}, and is now referred to as the physics-informed neural networks (PINNs) \cite{raissi2019physics}, which utilize the high expressibility of neural networks to approximate the solution of a differential equation. The PINNs model incorporates the governing equation into the network loss function, thereby transforming the problem of inferring solutions and identifying the unknown parameters of PDEs into an optimization problem of the loss function, which means it can solve the forward and inverse problems simultaneously. These characteristics have made it a focus of attention in the field of scientific machine learning \cite{chen2020physics,rasht2022physics}.

Meanwhile, numerous issues of PINNs have been identified, such as their inability to converge to the right solution \cite{krishnapriyan2021characterizing, wang2021understanding, wang2022and, de2023operator}, the high time burden of training \cite{bihlo2024improving, li2025improving}, the violation of physical principles \cite{hernandez2021structure}, etc. Especially towards the last issue, numerous efforts have been made by either inserting explicit physical laws, like the mass/momentum/energy conservation and entropy dissipation, into the loss function, or reformulating the differential equations into a more physical meaning structure to make the learning procedure more stable and the learned results more accurate \cite{jagtap2020conservative, cardoso2025exactly}. 

A systematic way to incorporate such physical structures is provided by the thermodynamic formalism, which offers a unified description of both conservative and dissipative dynamical systems. Conservative systems can be equivalently formulated through Newtonian, Lagrangian, and Hamiltonian mechanics \cite{greydanus2019hamiltonian, cranmer2020lagrangian, chu2024structure}, emphasizing force balance, variational principles, and energy-preserving symplectic structures, respectively. However, many realistic systems are inherently dissipative and require a thermodynamically consistent treatment, which is captured by formalisms such as the Onsager variational principle \cite{yu2021onsagernet}, Classical and Extended Irreversible Thermodynamics (CIT/EIT), the GENERIC framework \cite{zhang2022gfinns}, and Conservation-Dissipation Formalism (CDF) \cite{peng2021recent}. These approaches describe irreversible dynamics through entropy production, dissipation potentials, or coupled reversible-irreversible structures, thereby imposing intrinsic constraints on admission system evolution. Embedding these thermodynamic structures into PINNs provides a principled mechanism to constrain the solution space and improve the consistency and reliability of simultaneously solving forward and inverse problems. 

However, these previous works are based solely on conservation laws or dissipative structures to solve the forward and inverse problems of differential equations. To the best of our knowledge, there is no quantitative assessment on the impact of combining the PINNs model with different thermodynamics formalism on the performance of solving forward and inverse problems simultaneously. This constitutes the main motivation for the current study.

The remainder of this paper is organized as follows. Section \ref{Methods} reviews the fundamentals of PINNs, and the detailed formulation of various proposed thermodynamic-informed neural networks, including Lagrangian and Hamiltonian mechanics informed neural networks for conservative systems, Onsager's variational principle, and extended irreversible thermodynamics informed neural networks for dissipative systems. Section \ref{Results} presents numerical experiments comparing the performance of different thermodynamic PINN variants, including the ideal mass-spring oscillator, simple pendulum, and double pendulum as representative conservative systems, as well as the damped pendulum, diffusion equation, and Fisher-Kolmogorov equation as the dissipative examples. Furthermore, we investigate the generalization capabilities of different thermodynamics formalism-informed PINNs models from the perspective of the loss landscape, since the primary distinction among these models lies in their loss functions. Finally, Section \ref{Conclusion} summarizes the main findings and discusses potential directions for future research.

\section{Methods}\label{Methods}

\subsection{Problem setup}
To evaluate the numerical advantages of embedding different forms of dynamic equations of the same physical system (including both non-dissipative and dissipative systems) into PINNs models, we constructed multiple models incorporating these thermodynamic formalisms into neural networks. Our comparative analyses aim to include both forward and inverse problems: i.e., the ability of the model to learn the solutions to the equations and to infer the unknown parameters in the equations. The unique framework of the PINNs model enables the simultaneous solution of both forward and inverse problems associated with a dynamic equation, given partial data and knowledge of the equations. This capability stands out as a key advantage of the PINNs model.

For a system of parameterized nonlinear differential equations of the general form
\begin{eqnarray}
    \label{setup_eq}
&&\boldsymbol{u}_t + \boldsymbol{\mathcal{F}}[\boldsymbol{u}, \boldsymbol{\lambda}] = 0,\quad \forall t\in [0, T],\forall\boldsymbol{x}\in \Omega,\\
&&\mathcal{I}[\boldsymbol{u}](0,\boldsymbol{x})=0,\quad \forall\boldsymbol{x}\in \Omega,\nonumber\\
&&\mathcal{B}[\boldsymbol{u}](t,\boldsymbol{x})=0,\quad \forall t\in[0,T],\forall \boldsymbol{x}\in\partial\Omega.\nonumber
\end{eqnarray}
where $\boldsymbol{u}(t, \boldsymbol{x})$ is the solution of the differential equation, $\boldsymbol{\mathcal{F}}[\cdot, \boldsymbol{\lambda}]$ is the nonlinear operator parameterized by $\boldsymbol{\lambda}$. $\mathcal{I}[\boldsymbol{u}]$ and $\mathcal{B}[\boldsymbol{u}]$ stand for general initial and boundary conditions compatible with the differential equations. $\Omega$ is the subset of $\mathbb{R}^d$, $\partial\Omega$ represents the boundary of the domain $\Omega$, and $\boldsymbol{\lambda}$ may be unknown. Many works \cite{tartakovsky2018learning, tartakovsky2020physics} hope to find the most suitable $\lambda$ under a given data set and call this type of problem an inverse problem. On the other hand, many works \cite{yu2022gradient, guo2022monte} hope to design the model to obtain a data-driven solution that conforms to the equation (\ref{setup_eq}) when there is some data and $\boldsymbol{\lambda}$ is known, and this type of question is referred to as the forward problem. In this paper, we do not carefully distinguish between forward problems and inverse problems. We hope that when $\boldsymbol{\lambda}$ is unknown, the design model can find the most suitable parameters for a given small dataset, and then can  solve the equation (\ref{setup_eq}) in a fast and accurate way.

\begin{figure}[htbp]
    \centering
    \includegraphics[width=0.8\linewidth]{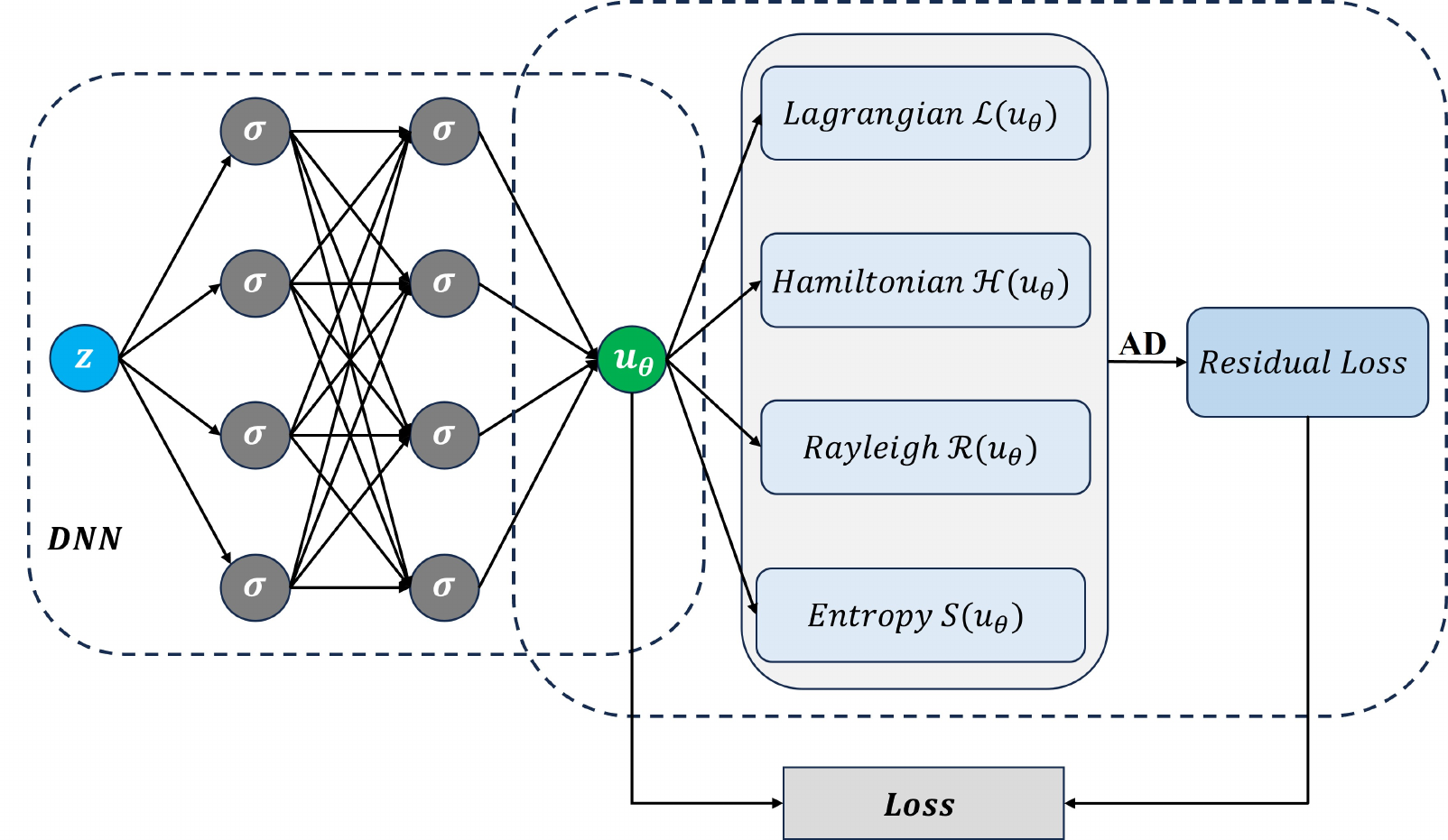}
    \caption{\textbf{The schematic diagram of the Lagrangian-mechanics, Hamiltonian-mechanics, Onsager's Variational Principle and Extended Irreversible Thermodynamics informed neural networks.}}
    \label{fig:placeholder}
\end{figure}

For the forward and inverse problems, to evaluate the accuracy of the learned solution $\hat{\boldsymbol{u}}(t,\boldsymbol{x})$, the $L^2$ relative error $\frac{\|\hat{\boldsymbol{u}}-\boldsymbol{u}\|_2}{\|\boldsymbol{u}\|_2}$ is used. While for the inverse problems, we use the relative error $\frac{|\hat{\boldsymbol{\lambda}}_i - \boldsymbol{\lambda}_i|}{\boldsymbol{\lambda}_i}$ is computed to evaluate the accuracy of the predicted coefficients $\hat{\boldsymbol{\lambda}}$. Due to randomness stemming from factors such as random sampling, network initialization, and optimization, each experiment is conducted by ten times. Subsequently, we compute the geometric mean of the errors for each case.

\subsection{Physics-informed Neural Networks}
Physics-informed Neural Networks(PINNs)\cite{raissi2019physics} eliminates the need for discretization of the solution domain, and the efficient optimization and prediction capabilities of neural networks are exploited. Following the original work, one can proceed by representing the unknown solution $\boldsymbol{u}(t, \boldsymbol{x})$ by a deep neural network $\boldsymbol{u}_{\theta}(t, \boldsymbol{x})$, where $\theta$ denotes all trainable parameters of the network, including the weights and biases. 
 Then, PINNs approximate the map between points in the spatio-temporal domain to the solution of the differential equations by minimizing the following composited loss function
 \begin{equation}
     Loss(\boldsymbol{\theta}) = \lambda_{data}Loss_{data}(\boldsymbol{\theta})+\lambda_{res}Loss_{res}(\boldsymbol{\theta})+\lambda_{ic}Loss_{ic}(\boldsymbol{\theta}) + \lambda_{bc}Loss_{bc}(\boldsymbol{\theta}).
 \end{equation}
The first term of the above loss function characterizes the difference between the neural network predictions and the labeled data $\boldsymbol{u}_{true}$ at those given points $(t_{data}^i, \boldsymbol{x}_{data}^i)$:
\begin{equation}
    Loss_{data}(\boldsymbol{\theta}) = \frac{1}{N_{data}}\mathop{\sum}\limits_{i=1}^{N_{data}}\big(\boldsymbol{u}_{\boldsymbol{\theta}}(t_{data}^i, \boldsymbol{x}_{data}^i) - \boldsymbol{u}_{true}(t_{data}^i, \boldsymbol{x}_{data}^i)\big)^2,
\end{equation}
where $t_{data}^i \in [0, T], \; \boldsymbol{x}_{data}^i \in \Omega$.
The second term of the loss function is the mean squared error due to the Residual of the differential equations:
\begin{equation}
    Loss_{res}(\boldsymbol{\theta}) = \frac{1}{N_{res}}\mathop{\sum}\limits_{i=1}^{N_{res}}\bigg(\frac{\partial \boldsymbol{u}_{\boldsymbol{\theta}}}{\partial t}(t_{res}^i, \boldsymbol{x}_{res}^i) - \mathcal{F}[\boldsymbol{u}_{\boldsymbol{\theta}}](t_{res}^i, \boldsymbol{x}_{res}^i)\bigg)^2,
\end{equation}
where $\frac{\partial \boldsymbol{u}_{\boldsymbol{\theta}}}{\partial t}(t_{res}^i, \boldsymbol{x}_{res}^i) - \mathcal{F}[\boldsymbol{u}_{\boldsymbol{\theta}}](t_{res}^i, \boldsymbol{x}_{res}^i)$ denotes the equation error predicted by the neural network at the given residual points $(t_{res}^i, \boldsymbol{x}_{res}^i)$, and $t_{res}^i \in [0, T], \; \boldsymbol{x}_{res}^i \in \Omega$.
The third term is the mean squared error of the initial condition:
\begin{equation}
    Loss_{ic}(\boldsymbol{\theta}) = \frac{1}{N_{ic}}\mathop{\sum}\limits_{i=1}^{N_{ic}}\big(\mathcal{I}[\boldsymbol{u}_{\boldsymbol{\theta}}](0, \boldsymbol{x}_{ic}^i)\big)^2, \quad \boldsymbol{x}_{ic}^i \in \Omega
\end{equation}
where $\mathcal{I}[\boldsymbol{u}_{\boldsymbol{\theta}}](0, \boldsymbol{x}_{ic}^i)$ represents the initial conditions predicted by the neural network at points $(0, \boldsymbol{x}_{ic}^i)$. And the last term shows the mean squared error about the boundary condition:
\begin{equation}
    Loss_{bc}(\boldsymbol{\theta}) = \frac{1}{N_{bc}}\mathop{\sum}\limits_{i=1}^{N_{bc}}\big(\mathcal{B}[\boldsymbol{u}_{\boldsymbol{\theta}}](t_{bc}^i, \boldsymbol{x}_{bc}^i)\big)^2, \quad t_{bc}^i\in [0, T], \quad \boldsymbol{x}_{bc}^i \in \partial \Omega
\end{equation}
where $\mathcal{B}[\boldsymbol{u}_{\boldsymbol{\theta}}](t_{bc}^i, \boldsymbol{x}_{bc}^i)$ represents the boundary condition of the PDEs predicted by the neural network at the boundary points $(t_{bc}^i, \boldsymbol{x}_{bc}^i)$. The hyperparameters  $\lambda_{data}$,  $\lambda_{res}$, $\lambda_{ic}$, and $\lambda_{bc}$ allow assigning a different learning rate to each loss term to balance their interplay during model training. For practical implementation, a finite number of labeled data points are prescribed on the initial and boundary conditions, and the corresponding data loss is evaluated jointly with labeled points sampled within the solution domain. This treatment is consistently adopted in all subsequent models based on various thermodynamic formulations.

\subsection{Lagrangian-Mechanics informed Neural Networks}

We first design the neural networks embedded with the Lagrangian theory of classical mechanics, called Lagrangian-Mechanics informed Neural Networks. Assuming that we have known the Lagrangian function of the system is $L=T-V$, which should be a function of the generalized coordinates $\boldsymbol{q}$ and $\boldsymbol{\dot{q}}$, abbreviated as $L = \mathcal{L}(\boldsymbol{q}, \dot{\boldsymbol{q}}, t)$. It is well-known that $L$ satisfies the Euler-Lagrange equation
\begin{equation}
    \frac{d}{dt}\frac{\partial L}{\partial \dot{\boldsymbol{q}}} - \frac{\partial L}{\partial \boldsymbol{q}} = 0.
\end{equation}
Consider embedding the Lagrangian function and Lagrangian equation into neural network models. First, use a neural network to approximate the function from time $t$ to the physical system coordinate $\boldsymbol{q}$. Then, we can use the automatic differentiation technique and the system's Lagrangian function $\mathcal{L}(\boldsymbol{q},\dot{\boldsymbol{q}},t)$ to automatically calculate the left-hand side term of the Lagrangian equation, which can then be embedded into the neural network model as the residual loss.

The loss function of the model primarily comprises the data loss, equation residual loss, and Lagrangian function loss, i.e.,
\begin{equation*}
    Loss = \lambda_{data}Loss_{data} + \lambda_{res}Loss_{res} + \lambda_{lag}Loss_{lag}.
\end{equation*}
Here $\lambda_{data},\lambda_{res},\lambda_{lag}$ are three adjustable weights. $Loss_{data}$ is the data loss, and let $\boldsymbol{\hat{q}}(t_{data}^i)$ and $\boldsymbol{\hat{\dot{q}}}(t_{data}^i)$ be the predicted results of the neural network, while $\boldsymbol{q}(t_{data}^i)$ and $\boldsymbol{\dot{q}}(t_{data}^i)$ be the real data at time $t_{data}^{i}$, one can get
\begin{equation}
    Loss_{data} = \frac{1}{N_{data}}\sum_{i=1}^{N_{data}}\Big[|\boldsymbol{\hat{q}}(t_{data}^i) - \boldsymbol{q}(t_{data}^i)|^2 + |\boldsymbol{\hat{\dot{q}}}(t_{data}^i) - \boldsymbol{\dot{q}}(t_{data}^i)|^2\Big].
\end{equation}
$Loss_{res}$ stands for the residual loss of the Euler-Lagrange equation. If we write $\boldsymbol{f} = \frac{d}{dt}\frac{\partial L}{\partial \dot{\boldsymbol{q}}} - \frac{\partial L}{\partial \boldsymbol{q}}$, then 
\begin{equation}
Loss_{res} = \frac{1}{N_{res}}\sum_{j=1}^{N_{res}}|\boldsymbol{f}(t_{res}^i)|^2.
\end{equation}
$Loss_{Lag}$ represents the constraint on the Lagrangian function, which can be calculated as 
\begin{equation}
Loss_{lag} = \frac{1}{N_{lag}}\sum_{i=1}^{N_{lag}}(\mathcal{L}(\hat{\boldsymbol{q}}(t_{lag}^i), \hat{\dot{\boldsymbol{q}}}(t_{lag}^i)) - L^i)^2.
\end{equation}
$\mathcal{L}(\hat{\boldsymbol{q}}(t_{lag}^i), \hat{\dot{\boldsymbol{q}}}(t_{lag}^i))$ is the Lagrangian quantity of the system predicted by the model at time $t_{lag}^i$, and $L^i$ is the real data of the Lagrangian at the same time point.

\subsection{Hamiltonian-Mechanics informed Neural Networks}

In addition to using the Lagrangian theory of classical mechanics to design PINNs models, we can also refer to the Hamiltonian theory to design corresponding PINNs models. Suppose that we have known the Hamiltonian function of the system $H = T+V$, which should be a function of generalized coordinates $\boldsymbol{q}$ and the generalized momentum $\boldsymbol{p}$ of the system, we shorten it as $H = \mathcal{H}(\boldsymbol{q}, \boldsymbol{p}, t)$. Furthermore, we know that the Hamiltonian function of the system satisfies the Hamiltonian canonical equation
\begin{equation}
    \boldsymbol{\dot{q}} = \frac{\partial H}{\partial\boldsymbol{p}}, \; \dot{\boldsymbol{p}} = -\frac{\partial H}{\partial \boldsymbol{q}}.
\end{equation}
To make full use of the existing physical information and solve the forward and inverse problems of the physical system at the same time, we consider using neural networks to approximate the mapping between $t$ and the generalized coordinates $\boldsymbol{q}$ and the generalized momentum $\boldsymbol{p}$ of the system. Furthermore, we can calculate the Hamiltonian of the system through the $\mathcal{H}(\boldsymbol{q}, \boldsymbol{p}, t)$, and automatically calculate each term in the Hamilton's canonical equation with the help of the technique of automatic differentiation. 

The loss function of this model is
\begin{equation*}
    Loss = \lambda_{data}Loss_{data} + \lambda_{res}Loss_{res}.
\end{equation*}
Let $\hat{\boldsymbol{q}}(t_{data}^i)$ and $\boldsymbol{\hat{p}}(t_{data}^i)$ be predicted results of the neural network, while $\boldsymbol{q}(t_{data}^i)$ and $\boldsymbol{p}(t_{data}^i)$ be the real data at time $t_{data}^{i}$. Then
\begin{equation}
    Loss_{data} = \frac{1}{N_{data}}\sum_{i=1}^{N_{data}} \Big[|\hat{\boldsymbol{q}}(t_{data}^i)-\boldsymbol{q}(t_{data}^i)|^2 + |\boldsymbol{\hat{p}}(t_{data}^i) - \boldsymbol{p}(t_{data}^i)|^2\Big].
\end{equation}
For the residual loss of Hamilton's canonical equation, one can write $\boldsymbol{f}_1 = \frac{d\boldsymbol{q}}{dt}-\frac{\partial H}{\partial \boldsymbol{p}}, \boldsymbol{f}_2 = \frac{d\boldsymbol{p}}{dt} + \frac{\partial H}{\partial \boldsymbol{q}}$, then 
\begin{equation}
Loss_{res} = \frac{1}{N_{res}}\sum_{j=1}^{N_{res}}\Big[|\boldsymbol{f}_1(t_{eq}^i)|^2 + |\boldsymbol{f}_2(t_{eq}^i)|^2\Big].
\end{equation}
Furthermore, if the physical system to be solved is a conserved system, we consider adding a loss term 
\begin{equation}
Loss_{cons} = \frac{1}{N_{cons}}\sum_{i=1}^{N_{cons}}|\frac{d\mathcal{H}}{dt}(\hat{\boldsymbol{q}}(t_{cons}^i), \hat{\boldsymbol{p}}(t_{cons}^i))|^2. 
\end{equation}
$\frac{d\mathcal{H}}{dt}(\hat{\boldsymbol{q}}(t_{cons}^i), \hat{\boldsymbol{p}}(t_{cons}^i))$ stands for the total derivative of the predicted Hamiltonian function with respect to time. Since the system is ideal, complete, and the active force is conservative, the energy of the physical system is conserved, and the time derivative of the system's Hamiltonian should be $0$, and the $Loss_{cons}$ adds this constraint to the loss function.

\subsection{OVP informed Neural Networks}
Dissipative physical systems, such as diffusion processes, chemical reaction networks, and non-equilibrium thermodynamics, are fundamentally governed by the Onsager's variational principle (OVP)\cite{onsager1931reciprocal,onsager1931reciprocal_2}, an extension of Rayleigh's principle of the least energy dissipation in Stokesian hydrodynamics. Given generalized coordinates $\boldsymbol{x}=(x_1, \cdots, x_m)$, where $m$ is the degree of freedom of the system, we can introduce the Rayleighian of the system:
\begin{equation}
    \mathcal{R} = \Phi + \dot{U} = \frac{1}{2}\mathop{\sum}\limits_{i,j}\xi_{ij}\dot{x}_i\dot{x}_j + \mathop{\sum}\limits_{i}\frac{\partial U}{\partial x_i}\dot{x}_i,
\end{equation}
where   $\Phi = \frac{1}{2}\mathop{\sum}\limits_{i,j}\xi_{ij}\dot{x}_i\dot{x}_j$
is called the dissipation function, and $\xi_{ij}$ is the friction coefficients, satisfying the Onsager reciprocal relation ($\xi_{ij}=\xi_{ji}$) and positive definiteness. $U(\boldsymbol{x})$ denotes the potential energy of the system.

The Onsager's variational principle states that the true time-evolution path of the system minimizes the Rayleighian, i.e.
\begin{equation}
    \frac{\delta R}{\delta \dot{x}_i} = \mathop{\sum}\limits_{j}\xi_{ij}\dot{x}_j + \frac{\partial U}{\partial x_i} = 0.
\end{equation}
The above equation is equivalent to the force balance equation
\begin{equation}
    \mathop{\sum}\limits_{j}\xi_{ij}\dot{x}_j = -\frac{\partial U}{\partial x_i}.
    \label{force_balance_eq}
\end{equation}
Let $(\xi^{-1})_{ij}$ be the inverse of the matrix $\xi_{ij}$, then equation (\ref{force_balance_eq}) gives a time-evolution equation for $x_i$ as
\begin{equation}
    \frac{dx_i}{dt} = -\mathop{\sum}\limits_{j}(\xi^{-1})_{ij}\frac{\partial U}{\partial x_j}.
\end{equation}
It can be shown that this variational structure guarantees physically consistent behaviors, such as monotonic energy dissipation, stability, and correct steady states. 

Here we develop a general framework called the Onsager-Variational-Principle Informed Neural Networks (OVP-PINNs), which integrates the Onsager's variational structure into physics-informed neural networks for learning dissipative dynamical systems. Suppose that we know the Rayleighian $\mathcal{R}(\boldsymbol{u}, \boldsymbol{v})$ of the system, a function of the system state $\boldsymbol{u}$ and the generalized velocity $\boldsymbol{v}$. By embedding the Rayleighian minimization condition $\frac{\partial \mathcal{R}}{\partial \boldsymbol{v}} = 0$ as a structural constraint in the loss function, the neural network is forced to learn dynamics that respect the correct thermodynamic geometry. The system states $\boldsymbol{u}$ and the generalized velocity $\boldsymbol{v}$ will be learned by the neural networks.

The total loss combines three terms
\begin{equation*}
    Loss = \lambda_{data}Loss_{data} +  \lambda_{res}Loss_{res}.
\end{equation*}
Let $\hat{\boldsymbol{u}}(\boldsymbol{x}_{data}^i, t_{data}^i)$ and $\hat{\boldsymbol{v}}(\boldsymbol{x}_{data}^i, t_{data}^{i})$ be the system states and generalized velocity predicted by the neural network, while $\boldsymbol{u}(\boldsymbol{x}_{data}^i,t_{data}^i)$ and $\boldsymbol{v}(\boldsymbol{x}_{data}^i,t_{data}^i)$ be the labeled data point at coordinate $(\boldsymbol{x}_{data}^i,t_{data}^{i})$. Then
\begin{equation}
    Loss_{data} = \frac{1}{N_{data}}\sum_{i=1}^{N_{data}}\Big[|(\hat{\boldsymbol{u}}-\boldsymbol{u})(\boldsymbol{x}_{data}^i,t_{data}^i)|^2 + |(\hat{\boldsymbol{v}} - \boldsymbol{v})(\boldsymbol{x}_{data}^i,t_{data}^i)|^2\Big].
\end{equation}
Meanwhile, the Onsager variational principle gives 
\begin{equation}
Loss_{res} = \frac{1}{N_{res}}\sum_{i=1}^{N_{res}}\bigg[\bigg|\frac{\delta\mathcal{R}}{\delta \hat{\boldsymbol{v}}}(\boldsymbol{x}_{res}^i, t_{res}^i)\bigg|^2\bigg]. 
\end{equation}
As a consequence, combining the Onsager variational principle with neural networks yields a physics-grounded, structure-preserving, and data-efficient learning framework for a wide range of dissipative systems.

\subsection{EIT informed Neural Network}

Extended Irreversible Thermodynamics (EIT) \cite{jou1988extended}, building on classical irreversible thermodynamics (CIT) \cite{prigogine1963introduction}, treats non-equilibrium flows such as heat flow and diffusion flow as independent thermodynamic state variables, and thus can effectively describe transient and nonlocal effects in nonequilibrium systems. Within the EIT frameworks, the entropy function 
\begin{equation}
    S = S(e, n, q, \Pi, \cdots),
\end{equation}
is written as a function of internal energy $e$, molecular density $n$, heat flux $q$, strain $\Pi$, etc., and satisfies the local entropy balance equation
\begin{equation}
\frac{\partial S}{\partial t} + \nabla\cdot\boldsymbol{J}_s=\sigma_s\geq 0,
\end{equation}
where $\boldsymbol{J}_s$ is the entropy flux and $\sigma_s$ is the entropy production rate. In CIT, entropy depends only on conserved variables such as energy density $e$ and particle number density $n$. Incorporating nonequilibrium fluxes into the entropy function allows EIT to capture memory, relaxation, and finite-speed propagation effects that are absent in CIT. 

Following the derivation procedure of EIT, it can be shown that each flow satisfies a relaxation-type equation. For example, the heat flow obeys the Cattaneo-Vernotte equation
\begin{equation}
    \tau_q \frac{dq}{dt}+q=-\kappa \nabla T,
\end{equation}
which guarantees the finite propagation speed and causality, in contrast to the classical heat-conduction equation with Fourier's laws.

To investigate its effectiveness when combined with deep learning, we incorporate EIT into physics-informed neural networks to learn dissipative dynamical systems directly from data and physical laws. Let $\boldsymbol{u}$ represent the system state together with its extended nonequilibrium variables, and let $\hat{\boldsymbol{u}}$ be its neural network approximation. The total loss function is defined as
\begin{equation}
    Loss = Loss_{data} + Loss_{res},
\end{equation}
where 
\begin{equation}
Loss_{data} = \frac{1}{N_{data}}\sum_{i=1}^{N_{data}}\big[|\hat{\boldsymbol{u}}(\boldsymbol{x}_{data}^{i}, t_{data}^i) - \boldsymbol{u}(\boldsymbol{x}_{data}^{i}, t_{data}^i)|^2\big]
\end{equation}
is the data loss term. Further assume the form of the local entropy function $S(e, n, q, \Pi, \cdots)$ is known, we can introduce the residue loss due to the enetropy balance equation:
\begin{equation}
    Loss_{ent} = \frac{1}{N_{ent}}\sum_{j=1}^{N_{ent}}\Big[\Big|\frac{\partial S}{\partial t}(\boldsymbol{x}_{ent}^j, t_{ent}^j) + \nabla\cdot\boldsymbol{J}_s(\boldsymbol{x}_{ent}^j, t_{ent}^j) - \sigma_s(\boldsymbol{x}_{ent}^j, t_{ent}^j)\Big|^2\Big].
\end{equation}
By enforcing the entropy balance law as a structural constraint, the resulting EIT-informed neural network is guided to learn dynamics that remain consistent with the fundamental principles of nonequilibrium thermodynamics.

To sum up, here we introduce several variants of PINNs by incorporating different thermodynamic formulations --  Hamiltonian and Lagrangian mechanics informed neural networks for modeling conservative dynamics, Onsager's variational principle and extended irreversible thermodynamics informed neural networks for dissipative dynamics, to be exact. A key distinction between them lies in residue loss (see Table \ref{table-1}). The models derived based on different thermodynamic formulations for the same problem may lead to dramatically distinguished performance in the procedure of machine learning, which is the key issue we hope to explore in the current study.

\begin{table}[htbp]
\centering
\caption{Comparison among different thermodynamics-informed neural networks.}
\label{tab:pinn_frameworks}
\renewcommand{\arraystretch}{1.35}
\setlength{\tabcolsep}{4pt}
\begin{tabular}{
p{1.7cm}
p{1.5cm}
p{1.5cm}
p{1.5cm}
p{3.8cm}
p{1.5cm}
}
\toprule
Framework 
& Theoretical Foundation 
& Pre-Knowledge 
& Neural Network 
& Residue Loss 
& Description \\
\midrule
NM-PINN 
& Newtonian Mechanics 
& $u(x,t)$ 
& $u_\theta(x,t)$ 
& $\left\lvert \dfrac{\partial u_\theta}{\partial t} - F[u_\theta] \right\rvert^2$ 
& Conservative or Dissipative Dynamics \\

LM-PINN 
& Lagrangian Mechanics 
& $L(q,\dot q,t)$ 
& $q_\theta(t)$ 
& $\left\lvert \dfrac{d}{dt}\!\left(\dfrac{\partial L}{\partial \dot q}\right) - \dfrac{\partial L}{\partial q} \right\rvert^2$ 
& Conservative Dynamics \\

HM-PINN 
& Hamiltonian Mechanics 
& $H(q,p,t)$ 
& $q_\theta(t)$ $ p_\theta(t)$ 
& $\left\lvert \dfrac{dq}{dt} - \dfrac{\partial H}{\partial p} \right\rvert^2 
 + \left\lvert \dfrac{dp}{dt} + \dfrac{\partial H}{\partial q} \right\rvert^2$ 
& Conservative Dynamics \\

OVP-PINN 
& Onsager's Variational Principle 
& $R[u,v]$ 
& $u_\theta(x,t)$ $ v_\theta(x,t)$ 
& $\left\lvert \dfrac{\delta R}{\delta v_\theta} \right\rvert^2$ 
& Dissipative Dynamics \\

EIT-PINN 
& Extended Irreversible Thermodynamics 
& $S[u,v]$ 
& $u_\theta(x,t)$ $ v_\theta(x,t)$ 
& $\left\lvert \dfrac{\partial S}{\partial t} + \nabla \cdot J_s - \sigma_s \right\rvert^2$ 
& Dissipative Dynamics \\
\bottomrule
\label{table-1}
\end{tabular}
\end{table}

\section{Results}\label{Results}
\subsection{Conservative Systems}
\subsubsection{Ideal Mass Spring}
Our first task is to model and solve the dynamics of an ideal mass-spring system \cite{goldstein2011classical}. Suppose there is an object with mass $m$ attached to the end of the ideal spring. Then we can calculate the acceleration driven by the elastic force according to Newton's second law of motion:
\begin{equation}
    m\ddot{q} = -kq,
\end{equation}
where $k$ is the elastic coefficient of the spring, and $q$ is the distance of the spring from its equilibrium position. For simplicity, we set $k = 1.0$ and $m=1.0$ in the experiment.

Meanwhile, this system can also be reformulated within the framework of Lagrangian and Hamiltonian mechanics. It is easy to show that the Lagrangian of this system reads
\begin{equation}
    L = T - V = \frac{1}{2}m\dot{q}^2 - \frac{1}{2}kq^2,
\end{equation}
and the Hamiltonian is
\begin{equation}
    H = T + V = \frac{1}{2}m\dot{q}^2 + \frac{1}{2}kq^2 = \frac{1}{2m}p^2 + \frac{1}{2}kq^2,
\end{equation}
where $p=m\dot{q}$ is the momentum of the system. The motion of the ideal mass spring is then governed by the Euler-Lagrange equation or Hamilton's canonical equations. 

\begin{figure}[htbp]
    \centering
    \includegraphics[width=1.0\linewidth]{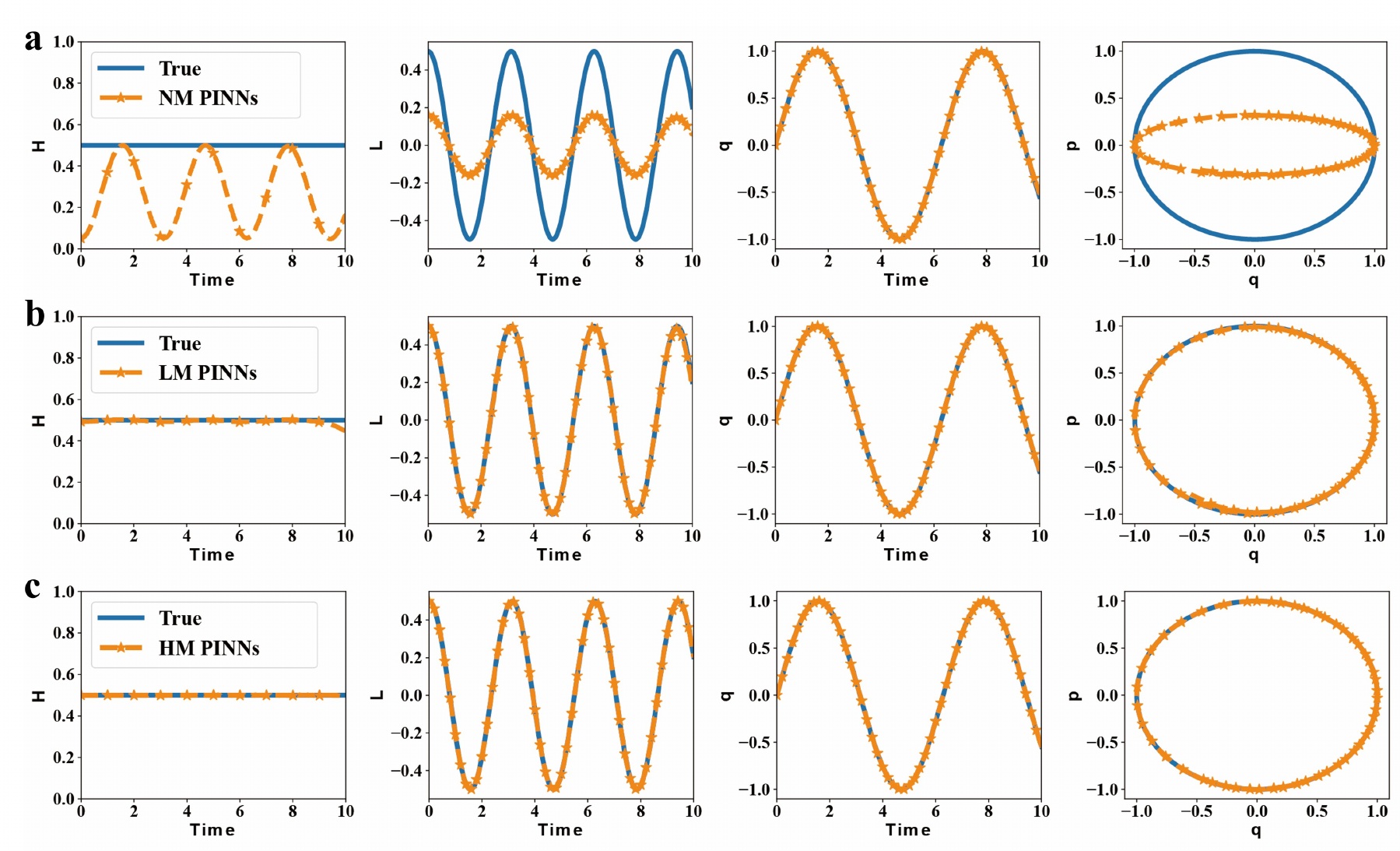}
    \caption{\textbf{Results for the forward problem of ideal mass spring.} The first column presents the predicted and true Hamiltonian, the second column shows the Lagrangian, the third column is the trajectory of the system, and the last column illustrates the phase diagram of $q$ and $p$. Panels (\textbf{a-c}) correspond to the results obtained using the NM-PINNs, LM-PINNs, and HM-PINNs, respectively.}
    \label{fig_ims}
\end{figure}

As a typical conserved system, it is natural to assess whether the solutions learned by NM-PINNs, LM-PINNs, and HM-PINNs preserve key physical quantities, such as the Lagrangian and Hamiltonian, in the forward problem. The first column of Figure \ref{fig_ims} (a-c) compares the Hamiltonian learned by the three models with the true value. Consistent with references \cite{cardoso2025exactly}, NM-PINNs fail to recover the conserved Hamiltonian. LM-PINNs achieve partial accuracy (MSE: $2.02\times 10^{-2}$), while the HM-PINNs, thanks to the addition of extra loss $Loss_{cons}$, can accurately learn the Hamiltonian with a high precision (MSE: $1.14\times 10^{-3}$). Notably, all three models can learn the accurate system state, as shown in the third column of Figure \ref{fig_ims} (a-c). We further compare the learned Lagrangian. NM-PINNs capture the periodic trend of the Lagrangian but exhibit substantial deviation from the exact value. In contrast, both LM-PINNs and HM-PINNs successfully recover the Lagrangian with high accuracy and maintain the correct periodic structure.

\begin{figure}[htbp]
    \centering
    \includegraphics[width=1.0\linewidth]{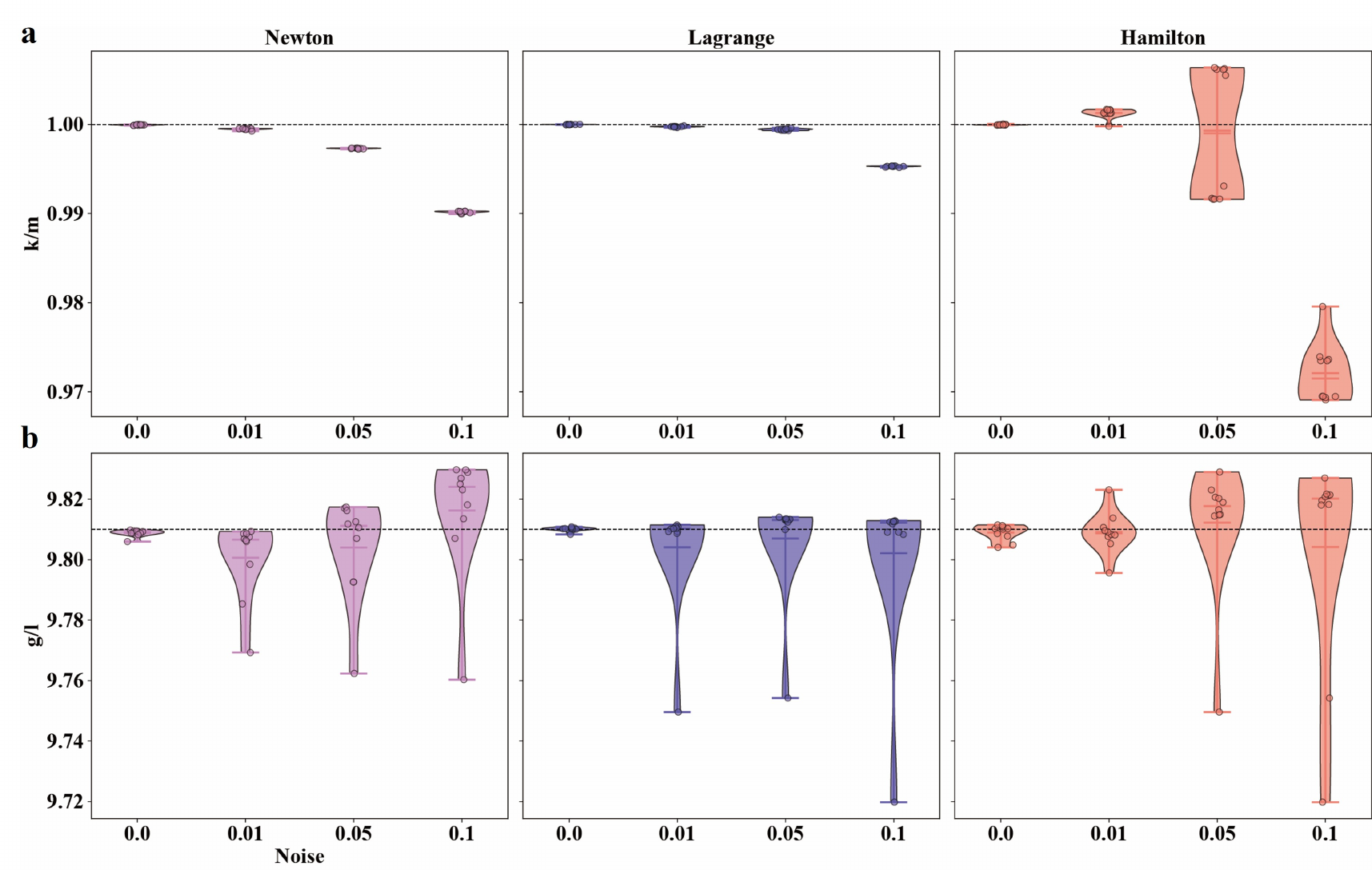}
    \caption{\textbf{Results for the inverse problem of ideal mass spring and ideal pendulum.} (\textbf{a}) For the ideal mass spring system, the learned distribution of $k/m$ obtained from the three models after ten random initializations under different noise levels. The black dashed line represents the ground truth value of $k/m$. (\textbf{b}) Corresponding results for ideal pendulum systems.}
    \label{fig_inverse_1}
\end{figure}

Different formulations of the governing equations used as residual loss also influence the performance on the inverse problem. Suppose both parameters $k$ and $m$ are unknown. Figure \ref{fig_inverse_1} (a) presents the distribution of $k/m$ inferred by each model under varied noise levels across ten training runs with random initialization, where the black dashed line represents the true $k/m$. All three models can achieve accurate identification in the noise-free setting. Under noisy conditions, however, LM-PINNs demonstrate the strongest robustness, maintaining a relative absolute error of approximately $4.72\times 10^{-1}$ even in the presence of $10\%$ noise. NM-PINNs rank second, whereas HM-PINNs exhibit the poorest performance. These differences arise from the structure of the residual equations. The Newton formulation involves second-order derivatives that strongly amplify noise. The Hamiltonian formulation depends on both position and momentum, and the increased number of noise-affected variables undermines model stability.

\subsubsection{Ideal Pendulum}

An ideal pendulum is a theoretical model consisting of a point mass $m$ suspended from a fixed point by a massless, inextensible string of length $l$, which swings under the influence of gravity without air resistance or friction. The nonlinear ordinary differential equation that governs the motion of the pendulum reads
\begin{equation}
    \ddot{q} + \frac{g}{l}\sin q = 0,
\end{equation}
where $q$ denotes the angular displacement and $g$ is the gravitational acceleration. For the ideal pendulum system, Lagrangian mechanics and Hamiltonian mechanics can also be used. For simplicity, we directly give the Lagrangian function of the system as
\begin{equation}
    L = T - V = \frac{1}{2}ml^2\dot{q}^2 - mgl(1-cosq),
\end{equation}
and the Hamiltonian function is
\begin{equation}
    H = T + V = \frac{1}{2ml^2}p^2 + mgl(1-cosq),
\end{equation}
where $p=ml^2\dot{q}$ is the momentum of the system. In this case, we set $m=1$, $g=9.81$ and $l=1.0$ in all experiments.

\begin{figure}[htbp]
    \centering
    \includegraphics[width=1.0\linewidth]{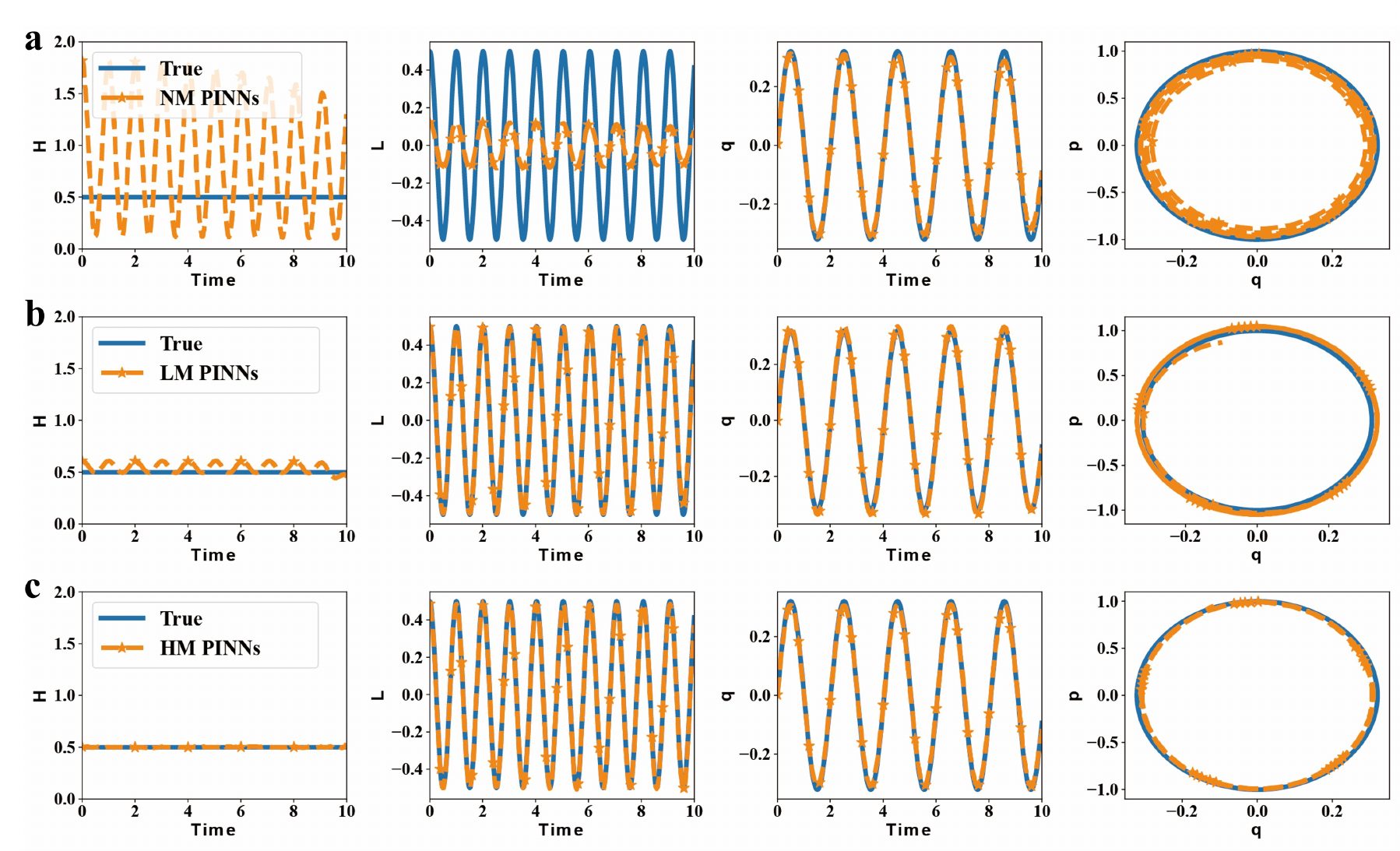}
    \caption{\textbf{Results for the forward problem of ideal pendulum.} The first column presents the predicted and true Hamiltonian, the second column shows the Lagrangian, the third column is the trajectory of the system, and the last column illustrates the phase diagram of $q$ and $p$. Panels (\textbf{a-c}) correspond to the results obtained using the NM-PINNs, LM-PINNs, and HM-PINNs, respectively.}
    \label{fig_ip}
\end{figure}

A central question is why residuals constructed based on different forms of the governing equations lead to such pronounced performance differences among the Newtonian, Lagrangian, and Hamiltonian models. In the ideal mass-spring system, NM-PINNs perform substantially worse than LM-PINNs and HM-PINNs in learning the phase portrait of $p$ and $q$ (see the fourth column of Figure \ref{fig_ims} (a-c)), which may be the main reason why NM-PINNs cannot accurately learn Lagrangian and Hamiltonian. Consistent with this behavior, the ideal pendulum, which shares a similar dynamic structure, exhibits the same trend, as shown in the fourth column of Figure \ref{fig_ip} (a-c). The Newton residual only constrains the second-order dynamics, without enforcing any structure on the first-order variables. In contrast, the Lagrangian and Hamiltonian formulations impose first-order structure-preserving constraints that directly regulate momentum and energy. The ability of HM-PINNs to simultaneously learn accurate Lagrangian and Hamiltonian representations comes with a trade-off. In the inverse problem (Figure \ref{fig_inverse_1} (b)), the mean value of $g/l$ inferred by HM-PINNs over ten random initializations under $10\%$ noise is comparable to that obtained by NM-PINNs, yielding relative absolute errors of $2.38\times10^{-1}$ and $1.71 \times 10^{-1}$, respectively. However, HM-PINNs exhibit a marked variance, indicating reduced stability despite similar average accuracy.

\subsubsection{Double pendulum}

A double pendulum is a classic example of a nonlinear dynamical system \cite{tabor1989chaos} consisting of two rigid rods connected in series, with the first pendulum attached to a fixed point and the second one attached to the end of the first. Each rod is assumed to be massless, and point masses $m_1$ and $m_2$ are located at their respective ends, with lengths $l_1$ and $l_2$. The configuration of the system can be described by two angular variables $\theta_1$ and $\theta_2$, representing the angular displacements from the vertical. For the sake of simplicity, we directly give the dynamics equation of the double pendulum system derived from the Newton mechanics
\begin{equation}
    \begin{split}
        \dot{\theta}_1 &= \omega_1,\\
        \dot{\theta}_2 &= \omega_2,\\
        \dot{\omega}_1 &= \frac{-g(2m_1+m_2)\sin\theta_1 - m_2g\sin(\theta_1-2\theta_2)-2\sin(\theta_1-\theta_2)m_2(\omega_2^2l_2+\omega_1^2l_1\cos(\theta_1-\theta_2))}{l_1(2m_1+m_2-m_2\cos(2\theta_1-\theta_2))},\\
        \dot{\omega}_2 &= \frac{2\sin(\theta_1-\theta_2)(\omega_1^2l_1(m_1+m_2)+g(m_1+m_2)\cos\theta_1+\omega_2^2l_2\cos(\theta_1-\theta_2))}{l_2(2m_1+m_2-m_2\cos(2\theta_1-2\theta_2))}.
    \end{split}
\end{equation}
The Lagrangian function of the double pendulum system is
\begin{equation}
    \begin{split}
        L & = \frac{m_1+m_2}{2}l_1^2\dot{\theta}_1^2+\frac{m_2}{2}l_2^2\dot{\theta}_2^2+m_2l_1l_2\dot{\theta}_1\dot{\theta}_2\cos(\theta_1-\theta_2)\\&+(m_1+m_2)gl_1\cos\theta_1+m_2gl_2\cos\theta_2.
    \end{split}
\end{equation}
Since the momentum $p_1 = \frac{\partial L}{\partial \dot{\theta}_1}, p_2 = \frac{\partial L}{\partial \dot{\theta}_2}$, one can obtain the Hamiltonian function as follows
\begin{equation}
    \begin{split}
        H &= \frac{l_2^2m_2p_1^2+l_1^2(m_1+m_2)p_2^2-2m_2l_1l_2p_1p_2\cos(\theta_1-\theta_2)}{2l_1^2l_2^2m_2[m_1+\sin^2(\theta_1-\theta_2)m_2]}\\&-m_2gl_2\cos\theta_2 - (m_1+m_2)gl_1\cos\theta_1.
    \end{split}
\end{equation}
For the uniformity of the experiment, we set $l_1=m_1=m_2=1.0, l_2=2.0$.

\begin{figure}[htbp]
    \centering
    \includegraphics[width=1.0\linewidth]{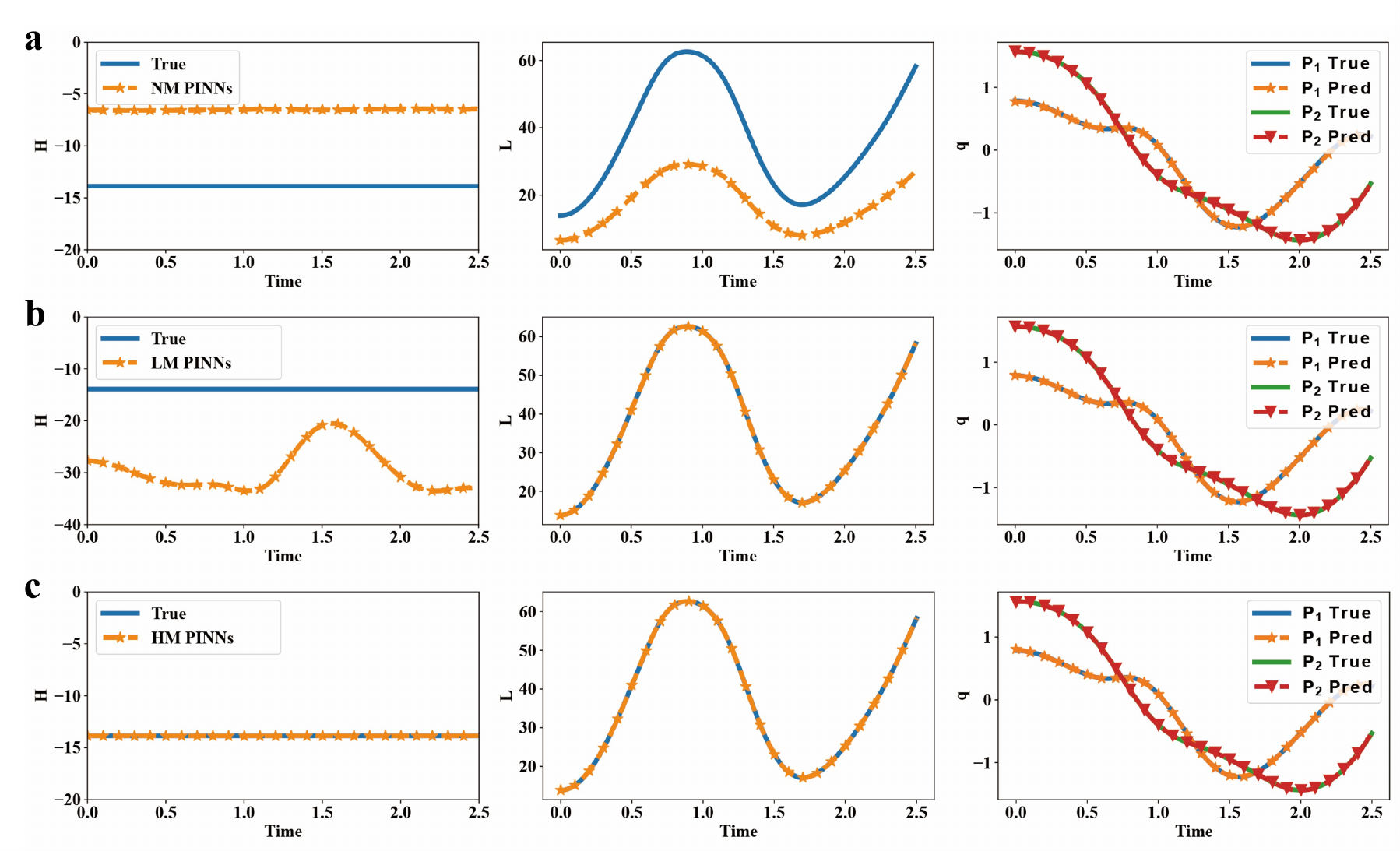}
    \caption{\textbf{Results for the forward problem of double pendulum.} The first column presents the predicted and true Hamiltonian, the second column shows the Lagrangian, and the third column is the trajectory of the system. Panels (\textbf{a-c}) correspond to the results obtained using the NM-PINNs, LM-PINNs, and HM-PINNs, respectively.}
    \label{fig_double}
\end{figure}

Although the double pendulum system is substantially more complex than the ideal pendulum case, the models exhibit broadly consistent behaviors in learning the Lagrangian and Hamiltonian. All three approaches recover accurate trajectories. However, Newton PINNs fail to capture the correct Lagrangian and Hamiltonian (Figure \ref{fig_double} (a)). Lagrange PINNs successfully learn the Lagrangian, but their predicted Hamiltonian shows substantial deviation from the ground truth(Figure \ref{fig_double} (b)). These findings further underscore that incorporating structured physical priors (especially Hamiltonian structures, see Figure \ref{fig_double} (c)) is crucial for enhancing the interpretability and reliability of PINN-based models.

\subsection{Dissipative Systems}
\subsubsection{Damped pendulum}

The damped pendulum provides a more realistic representation of the pendulum dynamics, as it accounts for energy dissipation that causes the oscillation amplitude to decay over time. Its equation of motion reads 
\begin{equation}
    \ddot{\theta} + \lambda \dot{\theta} + \beta^2 \theta = 0, \; \theta(t=0) = \theta_0,
\end{equation}
where $\theta$ is the angular displacement, $\lambda$ is the damping coefficient, and $\beta$ is the natural frequency in the absence of damping. The term $\lambda \dot{\theta}$ models the mechanism through which mechanical energy is gradually dissipated. The above equation will be used for constructing NM-PINNs. In this example, we set $\lambda=0.2, \beta=\sqrt{9.81}=3.1321$.

Alternatively, considering the structural equation $\dot{\theta} = \omega$, we can introduce the dissipation function $\Phi = \frac{1}{2}\lambda\omega^2$ and the potential function $U=\frac{1}{2}(\beta^2\theta^2+\omega^2)$, which leads to the Rayleighian
\begin{equation}
    R = \Phi+\dot{U} = \frac{1}{2}\lambda\omega^2 + \beta^2 \theta\dot{\theta} + \omega\dot{\omega}.
\end{equation}
Using the Onsager variational principle, one can get
\begin{equation}
    \frac{\delta R}{\delta \omega} = \lambda\omega + \beta^2\theta + \dot{\omega} = 0.
\end{equation}
Along with the structural equation $\dot{\theta} = \omega$, we arrive at the full model used for OVP-PINNs.

To formulate the dynamics from a thermodynamic perspective, consider the entropy function $S = S(\theta,\omega)=-\frac{1}{2}\beta^2\theta^2 - \frac{1}{2}\omega^2$, whose time derivative gives
\begin{equation}
\frac{dS}{dt}=\frac{\partial S}{\partial\theta}\frac{d\theta}{dt}+\frac{\partial S}{\partial\omega}\frac{d\omega}{dt}=-\beta^2\theta\frac{d\theta}{dt}-\omega\frac{d\omega}{dt}.
\end{equation}
By imposing constraint $\dot{\theta}=\omega$ and the constitutive equation $\dot{\omega}+ \beta^2\theta=-\lambda\omega$, the entropy balance equation leads to a non-negative entropy production rate
\begin{equation}
\frac{dS}{dt}=\sigma_s=\lambda\omega^2\geq0.
\end{equation}
in accordance with the second law of thermodynamics, as long as $\lambda\geq0$. The structural equation $\dot{\theta} = \omega$, the constitutive equation, as well as the entropy balance equation will all be included into the residue loss of EIT-PINNs.

\begin{table}[htbp]
    \centering
    \begin{tabular}{cccc}
        \toprule
        Noise & Model & $\lambda$ mean $\pm$ std & $\beta$ mean $\pm$ std (no outliers)\\
        \midrule
        & NM-PINNs & $0.2000 \pm 4.8021\times10^{-5}$ & $3.1321\pm2.2490\times10^{-5}$\\
        0\% & OVP-PINNs & $0.2000\pm3.9388\times10^{-5}$ & $3.1321\pm6.8259\times10^{-6}$\\
        & EIT-PINNs & $0.2000\pm 9.3730\times10^{-5}$ & $3.1273\pm6.2342\times10^{-3}$\\
        \midrule
        & NM-PINNs & $0.2002 \pm 1.9835\times10^{-4}$ & $3.1322\pm1.3322\times10^{-5}$\\
        1\% & OVP-PINNs & $0.2001\pm3.5251\times10^{-5}$ & $3.1324\pm9.3518\times10^{-6}$\\
        & EIT-PINNs & $0.2003\pm 1.3328\times10^{-4}$ & $3.1253\pm4.7730\times10^{-3}$\\
        \midrule
        & NM-PINNs & $0.1994 \pm 1.0121\times10^{-4}$ & $3.1325\pm1.3920\times10^{-5}$\\
        5\% & OVP-PINNs & $0.1996\pm3.3281\times10^{-5}$ & $3.1324\pm1.2373\times10^{-5}$\\
        & EIT-PINNs & $0.1977\pm 1.7854\times10^{-4}$ & $3.1198\pm5.0098\times10^{-3}$\\
        \midrule
        & NM-PINNs & $0.1989 \pm 8.8869\times10^{-5}$ & $3.1282\pm7.2125\times10^{-6}$\\
        10\% & OVP-PINNs & $0.2004\pm4.1065\times10^{-5}$ & $3.1319\pm4.5269\times10^{-5}$\\
        & EIT-PINNs & $0.1996\pm 6.1039\times10^{-4}$ & $3.1738\pm 4.2032\times10^{-3}$\\
        \bottomrule
    \end{tabular}
    \caption{\textbf{Unknown parameter prediction results of damped pendulum system under different noise intensities.} Ten random initialization simulations were performed for each model, and all results are averaged over ten simulations. "no outliers" excludes sign-flipped $\beta$ values caused by the $\beta^2$-based residual loss.}
    \label{tab_dp}
\end{table}

\begin{figure}[htbp]
    \centering
    \includegraphics[width=1.0\linewidth]{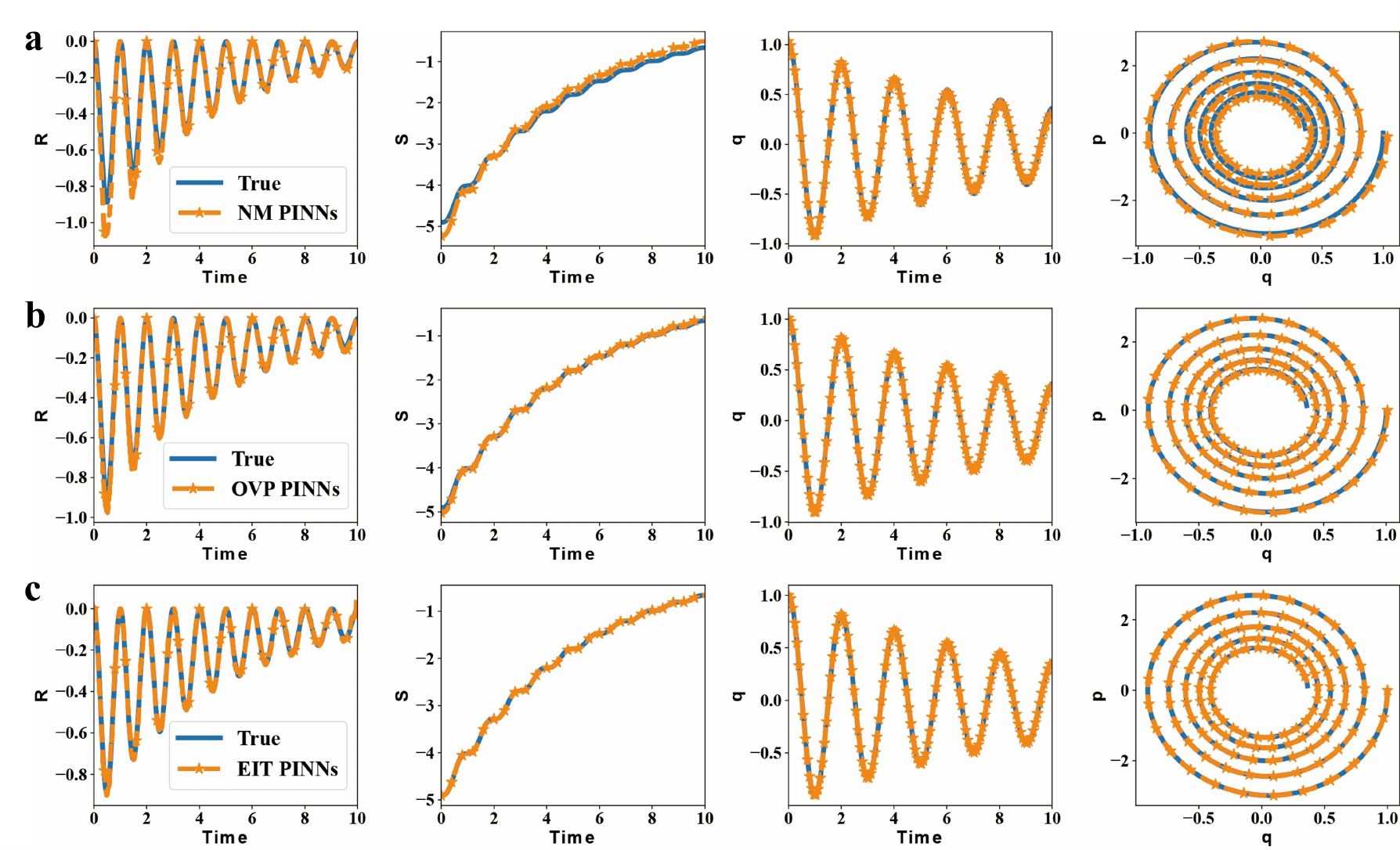}
    \caption{\textbf{Results for the forward problem of damped pendulum.} The first column presents the predicted and true Rayleighian, the second column shows the entropy function $S$, the third column is the trajectory of the system, and the last column illustrates the phase diagram of $q$ and $p$. Panels (\textbf{a-c}) correspond to the results obtained using the NM-PINNs, OVP-PINNs, and EIT-PINNs, respectively.}
    \label{fig_damp}
\end{figure}

As a prototypical dissipative system, we investigate the influence of using Newtonian, Onsager variational principle, and EIT form equation as residual loss on both forward and inverse problems. NM-PINNs, OVP-PINNs, and EIT-PINNs can all successfully reconstruct the system state trajectory (see the third and fourth columns of Figure \ref{fig_damp} (a-c)). However, pronounced differences emerge in their ability to learn the Rayleighian and entropy (first and second columns of Figure \ref{fig_damp} (a-c)). Specifically, the Rayleighian learned by the NM-PINNs model shows a noticeable bias at the early time (MSE: $1.59\times10^{-1}$), while OVP-PINNs and EIT-PINNs maintain high accuracy, with MSEs of $6.08\times10^{-2}$ and $3.89\times10^{-2}$, respectively. Moreover, NM-PINNs exhibit the largest error in recovering the entropy function (MSE: $5.75\times10^{-2}$), followed by OVP-PINNs (MSE: $1.33\times10^{-2}$). In contrast, EIT-PINNs achieve higi-precision recovery of the entropy prediction with an MSE of $1.47\times10^{-4}$. One notable advantage of OVP-PINNs emerges in inverse problems, particularly in the identification of $\beta$. Under $10\%$ noise, OVP-PINNs exhibit superior stability compared to both NM-PINNs and EIT-PINNs, substantially reducing the incidence of incorrect estimations (see Table \ref{tab_dp}).

\subsubsection{Diffusion equation}

The 2d diffusion equation describes the spatiotemporal evolution of a scalar field, such as temperature, concentration, or probability density, under the effect of diffusive transport. It is written as
\begin{equation}
    \frac{\partial u}{\partial t} = D_x \frac{\partial^2u}{\partial x^2} + D_y\frac{\partial^2 u}{\partial y^2}, \; \forall(x, y)\in \Omega, \; \forall t\in [0, T].
\end{equation}
where $u(x,y,t)$ denotes the field of interest, and $D_x, D_y > 0$ is the diffusion coefficients along the two spatial directions. In this case, we consider the Cauchy problem of the 2d diffusion equation and choose the initial condition to be a superposition of two anisotropic Gaussian kernels where the Gaussian centers are located at $(x_1, y_1)=(0.5, 0.5)$ and $(x_2, y_2)=(-0.5, -0.5)$. The diffusion coefficients used to generate the reference solution are $D_x=0.2$ and $D_y=0.5$.

To utilize Onsager's variation principle, we start with the continuity equation
\begin{equation}
\frac{\partial u}{\partial t} + \frac{\partial}{\partial x}(u v_1) + \frac{\partial}{\partial y}(u v_2) = 0.  
\end{equation}
Considering the dissipation function $\Phi=\frac{1}{2}\int_{\Omega}dxdy u\big[D_x^{-1}v_1^2 + D_y^{-1}v_2^2\big]$, and the potential energy function $U = \int_\Omega dxdy (u\ln u-u)$, we obtain the Rayleighian as 
\begin{equation}
    R = \Phi + \dot{U} = \int_\Omega \Big[\frac{1}{2}u\Big(D_x^{-1}v_1^2 + D_y^{-1}v_2^2\Big) + \frac{\partial u}{\partial t}\ln u\Big]dxdy,
\end{equation}
Then the variation of Rayleighian with respect to $v_1,v_2$ yields two constitutive relations for the velocities
\begin{equation}
    \begin{split}
        \frac{\delta R}{\delta v_1} &= D_x^{-1}uv_1 + \frac{\partial u}{\partial x} = 0,\\
        \frac{\delta R}{\delta v_2} &= D_y^{-1}uv_2 + \frac{\partial u}{\partial y}= 0,
    \end{split}
\end{equation}
by using integration by parts and infinite boundary conditions. The continuity equation together with two constitutive relations will be used for constructing the residue loss of OVP-PINNs.

The diffusion equation also fits into the EIT framework. The starting point is still the continuity equation $\frac{\partial u}{\partial t} + \frac{\partial}{\partial x}(u v_1) + \frac{\partial}{\partial y}(u v_2) = 0$. Define the local entropy function as $S[u]=-(u\ln u-u)$, then the local entropy balance equation reads
\begin{eqnarray}
&&\frac{\partial S}{\partial t}=\frac{\partial u}{\partial t}\frac{dS}{du}=\bigg[\frac{\partial}{\partial x}(u v_1) + \frac{\partial}{\partial y}(u v_2)\bigg]\ln u\nonumber\\
&=&\bigg[\frac{\partial}{\partial x}(v_1u\ln u) + \frac{\partial}{\partial y}(v_2u\ln u)\bigg]-u\bigg[v_1\frac{\partial}{\partial x}(\ln u) + v_2\frac{\partial}{\partial y}(\ln u)\bigg].
\end{eqnarray}
In the above equation, the first term is recognized as the local  entropy flux, while the second term denotes the local entropy production rate. To keep the non-negativity of the second term, it is natural to set 
\begin{equation}
    \begin{split}
        v_1&=-D_x\frac{\partial}{\partial x}(\ln u),\\
        v_2&=-D_y\frac{\partial}{\partial y}(\ln u).
    \end{split}
\end{equation}
It is clearly seen that EIT gives the same constitutive relations as those of OVP in the current case. Now, the local entropy production rate becomes $\sigma_s=u(D_xv_1^2+D_yv_2^2)$.

\begin{figure}[htbp]
    \centering
    \includegraphics[width=1.0\linewidth]{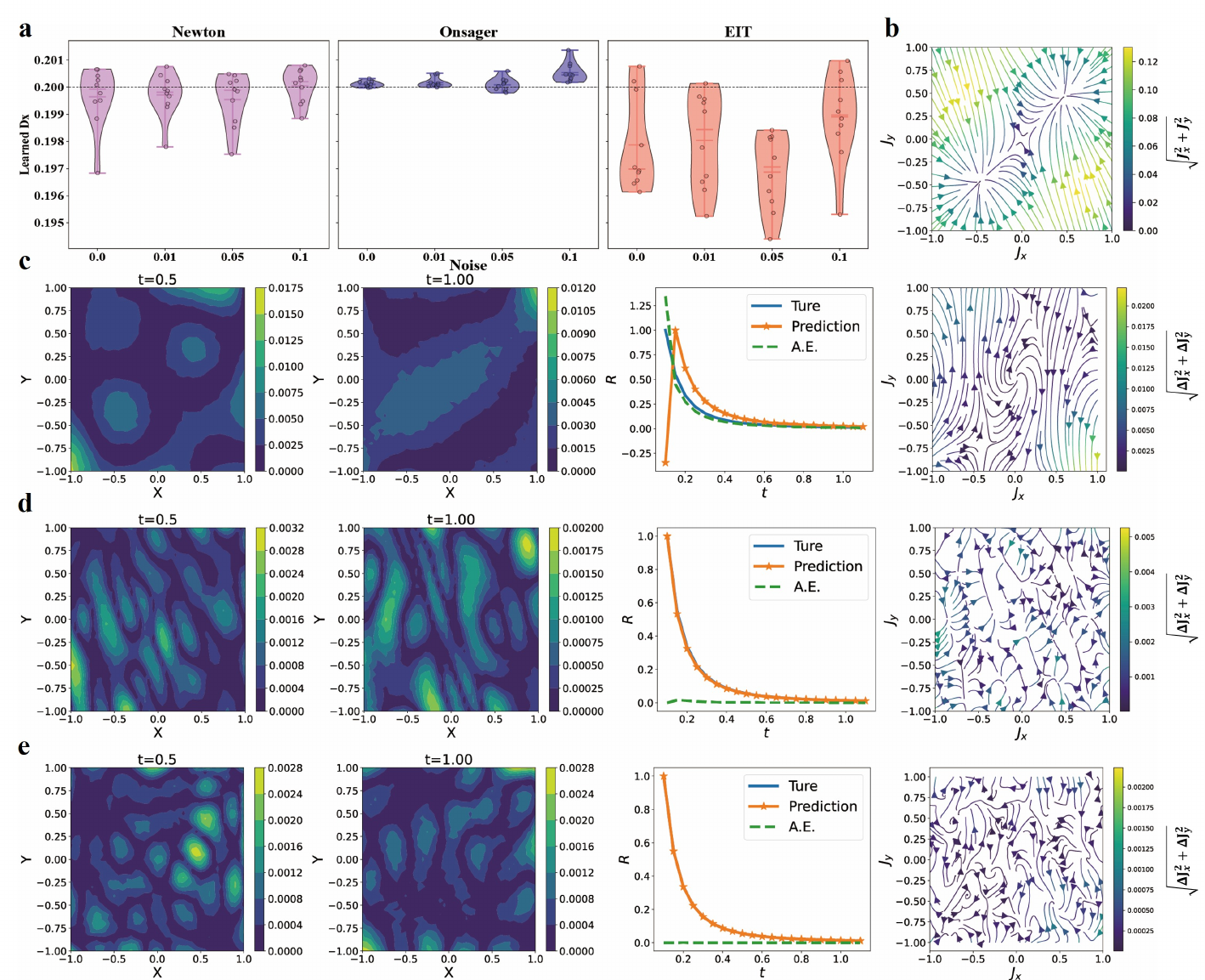}
    \caption{\textbf{Results for the forward problem of diffusion equation.} (\textbf{a}) The learned distribution of $D_x$ from the three models after ten random initializations under different noise levels. (\textbf{b}) The reference entropy flux at $t=0.5$. (\textbf{c-e}) Results obtained using NM-PINNs, OVP-PINNs, and EIT-PINNs, respectively. Each panel shows the point-wise absolute error at $t=0.5$ and $t=1.0$ (first two columns), the predicted normalized Rayleighian (third column), and the difference field between true and predicted entropy fluxes at $t=0.5$ (fourth column).}
    \label{fig_diffusion} 
\end{figure}

As a typical dissipative system, we aim to further compare the performance of NM-PINNs, OVP-PINNs, and EIT-PINNs on both forward and inverse problems. Figure $\ref{fig_diffusion}$ (a) presents the distribution of $D_x$ inferred by the three models under different noise intensities, each evaluated over ten random initializations. The results indicate that OVP-PINNs exhibit markedly stronger noise robustness than NM-PINNs and EIT-PINNs, as reflected by a significantly smaller variance. For the forward problem, all three models are capable of learning relative variance solutions, as shown in the first and second columns of Figure $\ref{fig_diffusion}$ (c-e), with corresponding RMSEs as $3.04\times 10^{-2}, 7.25\times10^{-3}$, and $2.88\times10^{-3}$, respectively. The primary distinctions among the models arise in their predictions of the Rayleighian and entropy flux. The third column of Figure $\ref{fig_diffusion}$ (c-e) compares the normalized Rayleighian inferred by each model with the ground truth value. NM-PINNs display substantial deviations at early times, including instances with incorrect signs. In contrast, although OVP-PINNs and EIT-PINNs also exhibit transient discrepancies, their predictions rapidly converge to correct Rayleighian and are significantly more accurate than those of NM-PINNs. The difference field of true and predicted entropy flux by the three models is shown in the fourth column of Figure $\ref{fig_diffusion}$ (c-e), while the analytically derived entropy flux is presented in Figure $\ref{fig_diffusion}$ (b). Among the three models, EIT-PINNs yields the smallest magnitude of entropy flux discrepancy and, more importantly, eliminates large scale coherent error structures. The residual differences exhibit only small scale, spatially uncorrelated patterns, indicating that the underlying entropy flux geometry is accurately captured. In contrast, NM-PINNs and OVP-PINNs suffer from systematic directional biases and localized structural mismatches.

\subsubsection{Fisher-Kolmogorov equation}

The Fisher-Kolmogorov (F-K) equation is a typical kind of reaction-diffusion equations, with the 1D general form
\begin{equation}
    \frac{\partial u}{\partial t} = D\frac{\partial^2 u}{\partial x^2} + \alpha u (1-u/K), \forall x \in \Omega, \forall t\in [0, T],
\end{equation}
where $u(x,t)$ represents the state variables of the spatiotemporal field (such as population density, chemical concentration, etc.), $D$ is the diffusion coefficient, $\alpha$ is the growth rate, and $K$ is the upper limit of the steady state (which we assume to be $1$). This equation was first proposed independently by Fisher \cite{fisher1937wave} and Kolmogorov-Petrovskii-Piskunov \cite{kolmogorov1study} to explain the spread and evolution of genes in a population, and it is currently widely used in ecology and brain science \cite{schafer2021bayesian,zhang2024discovering}.

Onsager's variational principle provides an insightful way to understand the F-K equation from a physical aspect. Start with the continuity equation 
\begin{equation}
\frac{\partial u}{\partial t} + \frac{\partial}{\partial x}(uv) = F(u),   
\end{equation}
where $F(u) =F^+(u)-F^-(u)$ with $F^+(u)=\alpha u$ and $F^-(u)=\alpha u^2$ denotes the source term due to chemical reactions. Considering the dissipation function $\Phi=\frac{1}{2}\int_\Omega dx(D^{-1} uv^2)$, and the potential energy function $U = \int_\Omega dx(u \ln u-u)$, we obtain the Rayleighian as 
\begin{eqnarray}
    R &=& \Phi + \dot{U}=\int_{\Omega}dx \bigg[\frac{1}{2}D^{-1}uv^2+\frac{\partial u}{\partial t}\ln u\bigg]\nonumber\nonumber\\
    &=&\int_{\Omega}dx \bigg[\frac{1}{2}D^{-1}uv^2+F(u)\ln u-\frac{\partial}{\partial x}(uv)\ln u\bigg]\nonumber\\
    .&=&\int_{\Omega}dx \bigg[\frac{1}{2}D^{-1}uv^2+F(u)\ln u+v\frac{\partial u}{\partial x}\bigg].
\end{eqnarray}
The last line is obtained by using integration by parts and infinite boundary conditions. Applying the Onsager variational principle yields the optimality conditions
\begin{equation}
\frac{\delta R}{\delta v} =D^{-1}uv+\frac{\partial u}{\partial x} = 0,\\
\end{equation}
Inserting it into the continuity equation, we recover the classical Fisher-KPP equation.

The F-K equation could be understood from a thermodynamic point of view too. With respect to the local entropy function $S(u)=-(u \ln u-u)$, the local entropy balance equation reads
\begin{eqnarray}
\frac{\partial S}{\partial t}&=&\frac{\partial u}{\partial t}\frac{dS}{du}=\frac{\partial(uv)}{\partial x}\ln u-[F^+(u)-F^-(u)]\ln u\nonumber\\
&=&\frac{\partial}{\partial x}(vu\ln u)-uv\frac{\partial}{\partial x}(\ln u)+[F^+(u)-F^-(u)]\ln\bigg(\frac{F^+(u)}{F^-(u)}\bigg).
\end{eqnarray}
By setting $v=-D\frac{\partial}{\partial x}(\ln u)$, it becomes clear that $\frac{\partial}{\partial x}(vu\ln u)$ denotes the local entropy flux, while the local entropy production rate includes two different contributions: $Du[\frac{\partial}{\partial x}(\ln u)]^2$ caused by particle diffusion and $[F^+(u)-F^-(u)]\ln[F^+(u)/F^-(u)]$ due to chemical reactions.

\begin{figure}[htbp]
    \centering
    \includegraphics[width=1.0\linewidth]{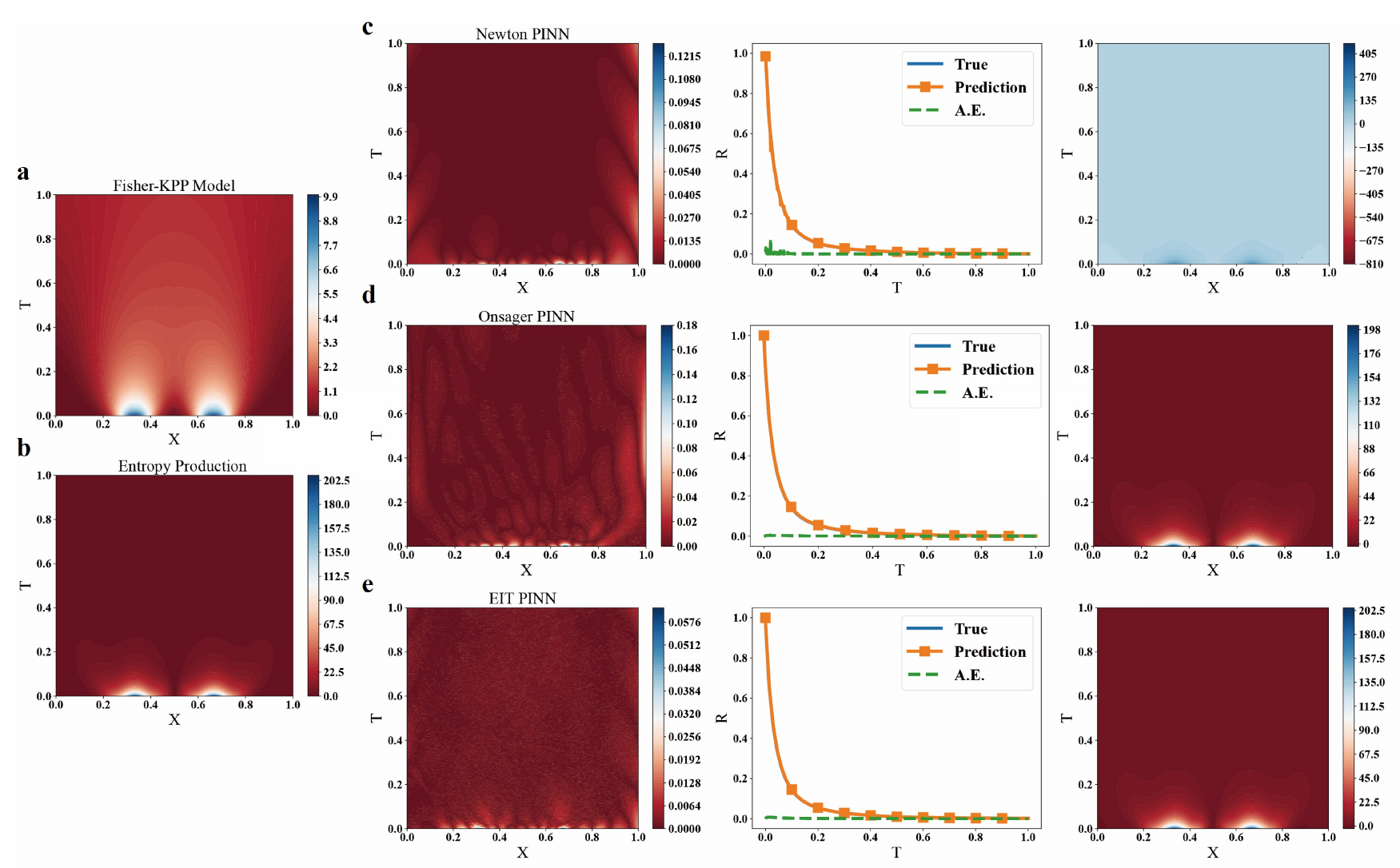}
    \caption{\textbf{Results for the forward problem of Fisher-Kolmogorov equation.} (\textbf{a}) Reference solution and (\textbf{b}) reference local entropy production rate for F-K equation. (\textbf{c-e}) Results obtained using NM-PINNs, OVP-PINNs, and EIT-PINNs, respectively. Each panel shows the point-wise absolute error of the predicted solution $u$ (first column), the predicted normalized Rayleighian (second column), and the predicted local entropy production rate (fourth column).}
    \label{fig_KPP} 
\end{figure}

In this example, EIT PINNs are trained using the local entropy balance equation derived from the entropy function $S[u]=-(u \ln u-u)$ as the residual loss, and their performance is compared with that of Newton PINNs and O-VPINNs. The first column of Figure $\ref{fig_KPP}$ (c-e) reports the absolute error between the solutions learned by the three models and the reference solution shown in Figure $\ref{fig_KPP}$ (a). EIT PINNs achieve the smallest maximum absolute error, and the RMSE values of the three models are $2.98\times 10^{-3}, 4.77\times10^{-3}$, and $1.39\times10^{-3}$, respectively. Regarding the learning of Rayleighian (second column of Figure $\ref{fig_KPP}$ (c-e)), NM-PINNs exhibit poor behaviors consistent with previous examples, characterized by pronounced errors and oscillations at early times. In this setting, possibly due to the adoption of a more general entropy function, EIT PINNs do not recover the Rayleighian as accurately as O-VPINNs. Nevertheless, owing to the incorporation of the local entropy balanced equation, EIT PINNs significantly outperform both Newton PINNs and O-VPINNs in learning the entropy production (third column of Figure $\ref{fig_KPP}$ (c-e)) and the RMSE associated with reference entropy production in Figure $\ref{fig_KPP}$ (b) is $2.84\times 10^{-2}$.

\subsection{Analysis by Loss Landscape}

Training neural networks amounts to minimizing a high-dimensional non-convex loss, which is NP-hard in theory \cite{blum1988training}, yet often tractable in practice. Empirical evidence shows that, despite the presence of many local minima, standard gradient-based methods frequently converge to solutions with similar performance, even though the data and labels are randomized before training\cite{zhang2021understanding}. However, such favorable behavior is not universal and strongly depends on network architecture, optimizer choice, and loss design. A long-standing belief holds that small-batch SGD tends to find "flat" minima with better generalization, whereas large batches converge to "sharp" minima \cite{hochreiter1997flat, keskar2016large, chaudhari2019entropy}, although this view has been challenged by subsequent studies \cite{dinh2017sharp, kawaguchi2017generalization}, and alternative training strategies have performed well even with large batch sizes \cite{hoffer2017train, de2017automated}. 

To address ambiguities in comparing loss landscapes, Li et al. \cite{li2018visualizing} introduced a filter-wise normalization technique that enables meaningful visualization and empirically supports the connection between flatter minima and better generalization. Specifically, for a loss function $\mathcal{L}(\boldsymbol{\theta})=\frac{1}{m}\sum_{i=1}^m l(\boldsymbol{x}_i, y_i;\boldsymbol{\theta})$, the local landscape around a reference point $\boldsymbol{\theta}^{*}$ is visualized via
\begin{equation}
    f(\alpha, \beta) = \mathcal{L}(\boldsymbol{\theta}^{*}+\alpha \boldsymbol{\delta} + \beta \boldsymbol{\mu}),
\end{equation}
where $\boldsymbol{\delta}$ and $\boldsymbol{\mu}$ are normalized random directions. Following \cite{li2018visualizing}, these directions are constructed by rescaling a Gaussian random vector $\boldsymbol{d}$ in a filter-wise manner,
\begin{equation}
    d_{ij} \longleftarrow \frac{d_{ij}}{\Vert d_{ij}\Vert} \Vert \theta_{ij}\Vert,
\end{equation}
thereby mitigating parameter-scaling effects and allowing fair comparisons across different layers and architectures.

\begin{figure}[htbp]
    \centering
    \includegraphics[width=1.0\linewidth]{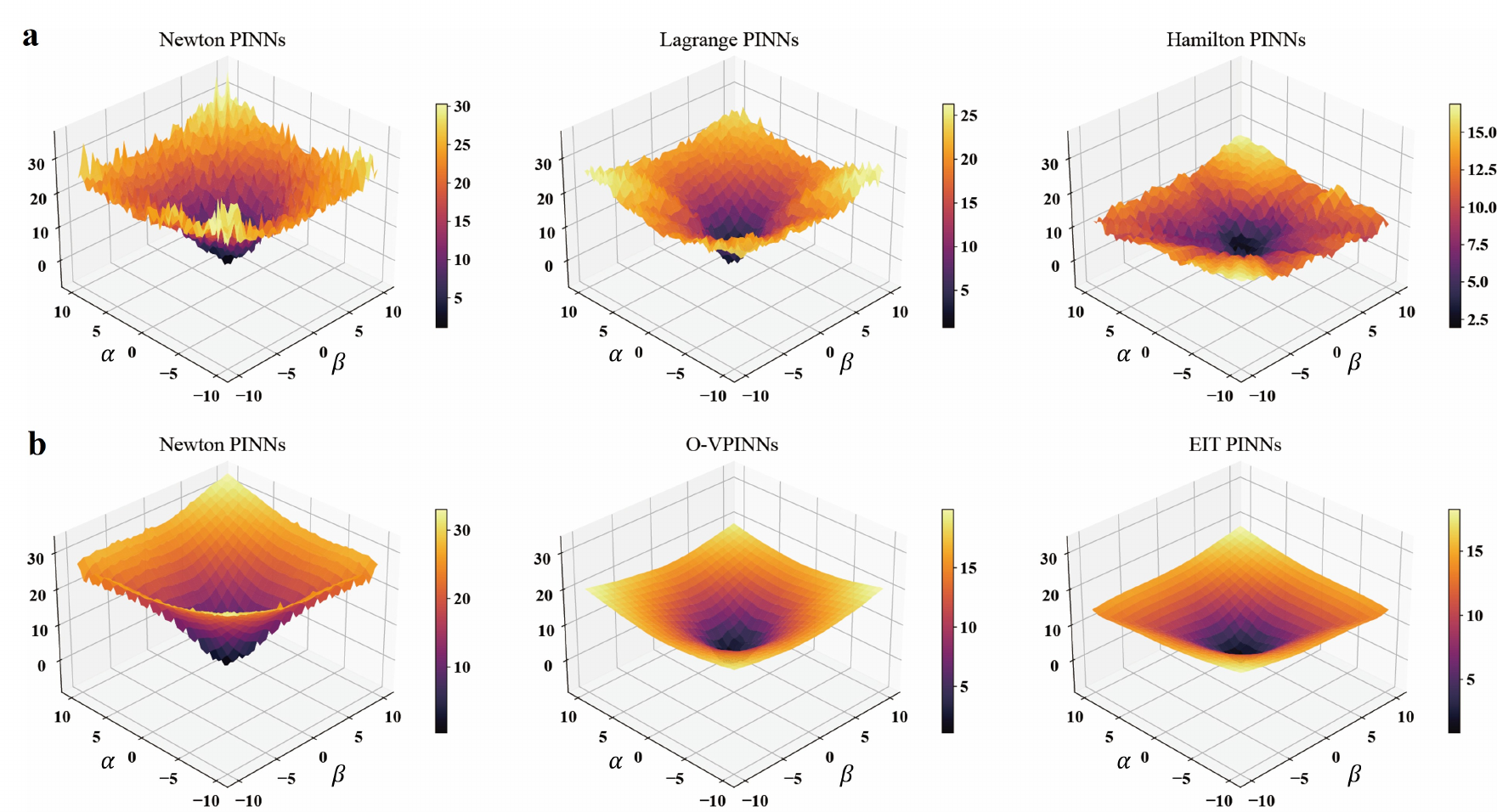}
    \caption{\textbf{Loss Landscape of ideal mass-spring system and diffusion equation.} (\textbf{a}) Loss Landscape of ideal mass-spring system, shown from left to right for NM-PINNs, LM-PINNs, and HM-PINNs. (\textbf{b}) Loss Landscape of diffusion equation, shown from left to right for NM-PINNs, OVP-PINNs, and EIT-PINNs.}
    \label{fig_lossLand} 
\end{figure}

Based on the method above, we visualized the loss landscape of different thermodynamic formalism-informed neural networks in the examples of mass-spring (Figure $\ref{fig_lossLand}$ (a)) and  diffusion (Figure $\ref{fig_lossLand}$ (b)). To quantify the flatness of the loss landscape, we calculated the average Frobenius norm
\begin{equation}
    \mathbb{E}[\Vert H \Vert^2_F], \; \Vert H\Vert_F^2 = \sum_{i,j}H_{ij}^2,\; H =\left(\begin{array}{cc}
         L_{xx} & L_{xy} \\
         L_{yx} & L_{yy}
    \end{array}\right),
\end{equation}
where $H$ is the Hessian matrix calculated based on the Loss Landscape. In Figure $\ref{fig_lossLand}$ (a), the average Frobenius norms calculated for NM-PINNs, LM-PINNs, and HM-PINNs are $214.64$, $43.47$, and $34.07$, respectively. In Figure $\ref{fig_lossLand}$ (b), the mean Frobenius norms corresponding to the loss landscape of NM-PINNs, OVP-PINNs, and EIT-PINNs are $15.25$, $1.28$, and $0.71$, respectively. This analysis clarifies why HM-PINNs and EIT-PINNs are able to learn additional physical quantities beyond the system trajectory. However, the aforementioned loss landscape analysis does not account for the superior performance of LM-PINNs and OVP-PINNs in inverse problems. In fact, accurate parameter recovery is governed by the global structure of the reduced loss with respect to physical parameters, rather than the local flatness of the full loss landscape. This issue is left to our future studies.

\section{Conclusion}\label{Conclusion}

In this work, we present a systematic comparison and analysis of thermodynamic structure informed neural networks, which is built on the general framework of PINNs by incorporating different thermodynamic formulations -- Newtonian, Lagrangian, and Hamiltonian mechanics for conservative systems, and the Onsager variational principle and extended irreversible thermodynamics for dissipative systems. Through comprehensive numerical experiments on a range of classical dynamical systems and partial differential equations (covering conservative and dissipative systems, ordinary and partial differential equations, and both forward and inverse problems), we quantitatively assess how different physical formulations used as residual constraints affect the model accuracy, physical consistency, noise robustness, and interpretability.

For conservative systems, including the ideal mass-spring oscillator, the simple pendulum, and the double pendulum, numerical experiments demonstrate that PINNs relying solely on Newtonian residuals can accurately reconstruct system trajectories but are markedly insufficient in learning key physical quantities such as the Lagrangian and Hamiltonian. Moreover, they exhibit instability in recovering the correct phase-space structure. In contrast, LM-PINNs offer clear advantages in parameter identification and noise robustness. HM-PINNs, by explicitly enforcing energy conservation, more faithfully preserve the Hamiltonian structure and energy invariants of the system, but their inverse problem stability decreases under high noise conditions. These findings clearly indicate that explicitly embedding structure-preserving physical formulations into PINNs is essential for enhancing physical consistency. 

For dissipative systems, such as damped pendulums, diffusion equations, and Fisher-Kolmogorov equations, NM-PINNs remain effective in reconstructing system states, but exhibit systematic biases when learning thermodynamic quantities such as the Rayleighian, entropy function, entropy flux, and entropy production. OVP-PINNs, by incorporating the Onsager variational principle, achieve enhanced stability and robustness in Rayleighian learning and parameter identification, particularly under noisy conditions. Furthermore, EIT-PINNs, through the explicit enforcement of entropy balance and entropy production constraints, are capable of recovering the entropy function and entropy flux in complex dissipative systems. While achieving comparable Rayleighian learning accuracy, EIT-PINNs significantly outperforms OVP-PINNs in terms of thermodynamic consistency and interpretability.

In summary, the results presented here provide clear empirical evidence to guide the systematic selection and design of appropriate thermodynamic formulations within the PINNs framework, and offer a foundation for integrating more general nonequilibrium thermodynamic structures, such as the GENERIC framework, with deep learning models, including neural operators.

\newpage

\backmatter

\bmhead{Code \& Data availability}

The source code and data for this project are available at \url{https://github.com/jay-mini/Thermodynamics-Formalism-informed-PINNs.git}.

\bmhead{Acknowledgements}

This work was supported by the National Key R\&D Program of China (Grant No. 2024YFA1011900), and the
Guangdong Provincial Key Laboratory of Mathematical and Neural Dynamical Systems (2024B1212010004). The authors thank Wuyue Yang for her helpful discussions.

\bmhead{Author Contributions}

Guojie Li: Investigation, Conceptualization, Methodology, Data curation, Formal analysis, Visualization, Writing-original draft. Liu Hong: Supervision, Funding Acquisition, Conceptualization, Project Administration, Writing-Review \& Editing. All authors reviewed the manuscript.

\bmhead{Competing Interests}

The authors declare that they have no conflict of interest.

\newpage
\bibliography{reference}


\end{document}